\documentclass[sigconf]{acmart}

\usepackage{balance}       
\usepackage{graphics}      
\usepackage[T1]{fontenc}   
\usepackage{txfonts}
\usepackage{mathptmx}
\usepackage{textcomp}
\usepackage[utf8]{inputenc}
\usepackage[justification=centering]{caption}
\usepackage{slashbox}
\usepackage[labelsep=quad,indention=10pt]{subfig}
\usepackage{microtype}        
\usepackage{ccicons}          
\usepackage{todonotes}
\usepackage{hyperref}

\copyrightyear{2019} 
\acmYear{2019} 
\setcopyright{acmcopyright}
\acmConference[ICTD '19]{THE TENTH INTERNATIONAL CONFERENCE ON INFORMATION AND COMMUNICATION TECHNOLOGIES AND DEVELOPMENT}{January 4--7, 2019}{Ahmedabad, India}
\acmPrice{15.00}
\acmDOI{10.1145/3287098.3287118}
\acmISBN{978-1-4503-6122-4/19/01}

\begin{document}
\title{Embedded CNN based vehicle classification and counting in non-laned road traffic}

\author{Mayank Singh Chauhan*}
\affiliation{%
  \institution{IIT Delhi}
}
\email{cs5160394@cse.iitd.ac.in}

\author{Arshdeep Singh*}
\affiliation{%
  \institution{IIT Delhi}
}
\email{arshdeep.cs516@cse.iitd.ac.in}

\author{Mansi Khemka*}
\affiliation{%
  \institution{Delhi Technological University}
}
\email{mansikhemka\_bt2k15@dtu.ac.in}

\author{Arneish Prateek}
\affiliation{%
  \institution{IIT Delhi}
}
\email{ch1140786@chemical.iitd.ac.in}

\author{Rijurekha Sen}
\affiliation{%
  \institution{IIT Delhi}
}
\email{riju@cse.iitd.ac.in}

\renewcommand{\shortauthors}{M. Chauhan et al.}

\begin{abstract}
Classifying and counting vehicles in road traffic has numerous applications in the transportation engineering domain. However, the wide variety of vehicles (two-wheelers, three-wheelers, cars, buses, trucks etc.) plying on roads of developing regions without any lane discipline, makes vehicle classification and counting a hard problem to automate. In this paper, we use state of the art Convolutional Neural Network (CNN) based object detection models and train them for multiple vehicle classes using data from Delhi roads. We get upto 75\% MAP on an 80-20 train-test split using 5562 video frames from four different locations. As robust network connectivity is scarce in developing regions for continuous video transmissions from the road to cloud servers, we also evaluate the latency, energy and hardware cost of embedded implementations of our CNN model based inferences.\\
* These authors have equal contributions.
\end{abstract}

%
%
\ccsdesc[500]{Computer systems organization~Embedded systems}

\keywords{Urban computing, traffic monitoring, deep neural network, computer vision, embedded computing}

\newcommand{\one}{front\_fisheye}
\newcommand{\two}{back\_nobus}
\newcommand{\three}{side\_highway}
\newcommand{\four}{back\_busroute}

\newcommand*{\origrightarrow}{}
\let\oldarrow\textrightarrow
\renewcommand*{\textrightarrow}{\fontfamily{cmr}\selectfont\origrightarrow}

\maketitle
\section{Introduction}
Traffic congestion and air pollution levels are becoming life threatening in developing region cities like Delhi and the National Capital Region (NCR). The local government is being forced to take concrete steps to make public transport better by gradually adding subway and bus infrastructure~\cite{metro-expansion}, albeit under budget constraints. Policy decisions like odd-even rules (vehicles with odd and even numbered plates are allowed on alternate days) are being tried to curb the number of private cars~\cite{oddeven, oddeven2, oddeven3}. A lot of times such policy decisions are met with angry protests from citizens in news and social media. In absence of data driven empirical analysis of the potential and actual impact of such urban transport policies, debates surrounding the policies often become political rhetoric. Building systems to gather and analyze transport, air quality and similar datasets is therefore necessary, for data driven policy debates.

This paper focuses on a particular kind of empirical measurement, namely counting and classification of vehicles and pedestrians from roadside cameras installed at intersections in Delhi-NCR. These numbers can be used in road infrastructure planning, e.g. in construction of signalized intersections, fly-overs, foot-bridges, underpasses, footpaths and bike lanes. Classified counts can also help in evaluating the effect of policies like odd-even, to see if private transport numbers go down during the policy enforcement period as expected. We discuss these and more motivational use cases of automated vehicle counting and classification in Section~\ref{sec-motivation}. We show how some of these use cases can benefit from empirical data, based on our dataset, in Section~\ref{sec-applications}.

Non-laned driving in developing regions with high heterogeneity of vehicles and pedestrians, make automated counting and classification a hard problem. This paper explores state of the art computer vision methods of CNN based object detection, to handle this task. CNN models need annotated datasets from the target domain for supervised learning. Annotated video frames from in-vehicle and roadside cameras are available for western traffic, and have recently been in high demand to train computer vision models for self-driving cars etc. 

We started our explorations with CNN models, available with weights trained on the Imagenet dataset~\cite{imagenet}, which has many classes of objects including vehicles. We further fine-tuned the model with existing annotated datasets of developed country traffic from PASCAL VOC~\cite{pascalvoc} and KITTI~\cite{kitti2}. However, the accuracies obtained with models trained with developed country traffic datasets, on test videos and images collected in Delhi-NCR, was very low (Mean Average Precision or MAP value in object detection was 0.01\% using fine-tuning with KITTI dataset and 0.58\% using fine-tuning with Pascal VOC).

We identified several differences between the annotated video and image datasets of western traffic and our traffic videos, that might cause the accuracy difference of training models on one dataset and testing on the other. Four-wheelers and motorbikes look similar across countries, but there are many vehicles in Delhi-NCR which look completely different from the western world (e.g. auto-rickshaws, e-rickshaws, cycle-rickshaws, trucks and buses). Secondly, our lack of lane discipline causes higher levels of occlusion, where a large vehicle like a bus is occluded by many smaller vehicles. Thirdly, our roads are not rectangular grid shaped as seen in developed country videos, but have different adhoc intersection designs, creating different views of the captured traffic flows. Finally, since self-driving cars is one of the main application focus in developed countries, many images and videos are captured from the view point of the driver. This view significantly differs from the view of traffic a road-side camera gets. On obtaining low accuracies with annotated datasets from developed regions as a combined effect of all these differences, we tried to find annotated datasets of non-laned developing region traffic to fine-tune our CNN models. Unfortunately, we could not find such datasets.

We therefore create such annotated datasets ourselves, as part of this paper. We collect videos from three different intersections and a highway in Delhi-NCR, in collaboration with Vehant Technologies~\cite{vehant} and Delhi Integrated Multimodal Transit System (DIMTS)~\cite{dimts}. We split the video into frames, manually annotate the different objects with bounding boxes and use this annotated dataset to train and test CNN based object detectors. Our annotated dataset comprises 5562 frames with 32088 total annotations, average number of annotations per frame being 6. In an 80-20 split of train and test data, our trained model achieves MAP values of upto 75\%. We describe our dataset in Section~\ref{sec-dataset} and CNN based object detector model training and testing in Section~\ref{sec-training}. 

The cameras from which we obtain data have either a fish-eye or a normal lens, and get a frontal, back or side view of the road traffic. These different kind of camera installations and lens configurations help us in evaluating how models trained on one annotated dataset perform on test set from the same camera vs. other installations. Our observations should empirically motivate the standardization of such hardware installation in future, to reduce the overhead of manual annotation of video frames and retraining of CNN model for each non-standard camera installation. Already fine-tuning of computer vision models, trained with videos and images of developed world traffic is needed. This is because we cannot change the kind of vehicles that ply on our roads, nor can we change our non-laned driving increasing occlusion, and also not the irregular intersection design different from regular grids in developed regions. But at least if differences due to camera positioning and angles can be minimized, some manual annotation and fine-tuning efforts can be reduced.

Vehicle counting and classification can be useful in two kind of applications. The first kind is delay tolerant, where processing can be done at any arbitrary latency after video capture. The computations in this case will affect long term policy like infrastructure planning or help in evaluating policy impact like that of odd-even rule. The second kind of applications require low latency real time processing. Here the computations can be used in catching speed violations based on vehicle class, or illegal use of roadways by heavy vehicles outside their allotted time slots. Low latency is needed to catch the violators and penalize them in real time. In this paper, we therefore also explore the prospect of real time inferences using our trained CNN models, especially using on-road embedded platforms.

Why is embedded processing interesting to explore in this context? Since broadband network connectivity across different road intersections and highways is not reliable in developing countries, transfer of video frames from the road to cloud servers for running computer vision models on them can become a bottleneck. We therefore evaluate embedded platforms on their ability to run inference tasks i.e. given a pre-trained CNN based object detection model and a video frame, whether the embedded platform can process the frame to give classified counts. We measure the latency incurred and energy drawn per inference task on three off-the-shelf embedded platforms (Nvidia Jetson TX2, Raspberry PI Model 3B and Intel Movidius Neural Compute Stick). Our evaluations in Section~\ref{sec-embedded} show the feasibility of embedded processing and also shows the cost-latency-energy trade-offs of particular hardware-software combinations.

Our trained models are available at\footnote{https://github.com/mansikhemka/Embedded-CNN-based-vehicle-classification-and-counting-in-non-laned-road-traffic/ for use by both computer vision and transportation researchers, and potentially also by government organizations working on road traffic measurement and management. Our annotated datasets will be available on request from academic and research institutions. This restriction is needed to control data privacy, as the videos have been captured on real roads and contain personally identifiable information (PII) like people's faces and vehicle number plates.} The annotated datasets will potentially be of interest to computer vision researchers, for designing and testing better CNN models for developing region traffic. The trained models and technical know-how of training the CNN models and running inferences on embedded platforms will potentially aid government organizations in data driven policy design and evaluation on road traffic measurement and management.

\begin{table*}[t!]
\begin{center}
\begin{tabular}{|l||*{5}{c|}}\hline
\backslashbox{Description}{Installation}
&\makebox{\one}&\makebox{\two}&\makebox{\three}&\makebox{\four}&\makebox{total}\\\hline\hline
Total Number of Annotated Images &620&1189&2754&999&5562\\\hline
Total Number of Annotations in all Images&9053&12588&4786&5661&32088\\\hline
Average Number of annotations per Image&~15&~11&~2&~6&~6\\\hline
Number of Train Images(80\%)&496&988&2248&800&4532\\\hline
Number of Test Images(20\%)&124&201&506&199&1030\\\hline
\end{tabular}
\caption{Image dataset and annotation description}
\vspace{-0.7cm}
\label{tab-dataset}
\end{center}
\end{table*}

\section{Motivation}
\label{sec-motivation}
Why is vehicle classification and counting useful? One use case for such classified counts is data driven infrastructure planning. Each vehicle class can carry a certain number of passengers, which is called Passenger Count Unit (PCU)~\cite{pcu}. PCU/hour is used to compute capacity of roads and if this capacity needs to be increased, flyovers, underpasses and road widening projects have to be undertaken after proper assessment. While this is the norm in developed countries, in developing countries infrastructure enhancement projects are often ridden in political rhetoric and controversy. A notable example has been civil society's vehement protests against local government's decision to build the Bengaluru steel flyover~\cite{flyover1, flyover2}. Such decisions and associated debates should be backed with empirical data of PCU measurements from the road, for which vehicle classification and counting as done in this paper is necessary. Other infrastructure like footbridges, footpaths and bike-lanes can also be planned based on counting actual numbers of pedestrians and cycles on the road.

A second use case is empirically evaluating the effect of urban transport policies. An example is the odd even policy piloted twice by the local government in Delhi-NCR in 2017~\cite{oddeven, oddeven2, oddeven3} to reduce number of vehicles, and subsequently fuel emissions to improve air quality index in the city. Again the policy was highly debated in news and social media. As shown by the researchers in~\cite{oddevenstudy}, most of these debates were driven by political leanings of the social media users, instead of empirical data. There were also controversial news reports of citizens resorting to renting cars with suitable license plates, and even opting to buy two cars with one odd and one even licence plate, to bypass the restrictions~\cite{oddevennoeffect1, oddevennoeffect2, oddevennoeffect3}. Actual numbers from the roads are necessary to quantify whether the government provided more public transport to ease commute during the pilot period, or whether the number of privately owned two-wheelers and four-wheelers actually went down in presence of people trying to fool the system. Automated vehicle counting and classification can directly provide these numbers of public and private vehicles for data driven policy audits. Also in connection to air quality improvement, different vehicle types are known to have different fuel emission properties~\cite{vehicleemissions}. Hence classified counts of vehicles will also be useful to correlate with air quality measurements.
 
Two other use cases were provided to us by DIMTS, who also shared their camera dataset. One was to detect buses so that DIMTS can see the arrival rates of buses at the point of monitoring. This can quantify the unpredictability of public transport arrival (do buses come every t minutes or is there a large variance in arrival times?). DIMTS is also instrumenting buses with GPS to get this information, but since some traffic cameras are already in place, they are interested to leverage that infrastructure for tracking buses till the GPS system comes up. The second use case is to detect heavy vehicles like trucks to penalize them if they drive outside their allowed hours. On the penalization side, different vehicles also have different speed limits\footnote{ https://en.wikipedia.org/wiki/Speed\_limits\_in\_India}, so vehicle classification is needed for speed violation detection as well (just detecting vehicle speed is not enough as each type has a different limit).

Finally, traffic management might benefit from more fine-grained information on vehicle types. Traffic density or queue length might be enough to better schedule the signals at intersections. But since there is significant heterogeneity in speeds of different vehicle types which take different times to clear the signals, whether signal scheduling should take into account more fine-grained information like classified vehicle counts, needs to be investigated.

\section{Related Work}
Using CNN based accurate computer vision methods, advanced forms of road and traffic related automation, e.g. self driving cars, is being investigated. Annotated image datasets to train the intelligent agent in self driving cars are being created as a result~\cite{trancos, kitti2, cityscapes}. Computer vision researchers across the world are designing and testing their CNN models on these datasets. These datasets are available for laned traffic of developed countries, where some vehicles like auto-rickshaw, e-rickshaw and cycle-rickshaws are absent, and some vehicles like trucks, buses and commercial vans look very different from those in developing region traffic. Also non-laned traffic leads to higher levels of occlusion. These differences led to very poor accuracy when we tried CNN models trained with annotated datasets of lane-based traffic, on test images from Delhi-NCR. We used an available CNN model with weights trained on the Imagenet dataset~\cite{imagenet}, which has many classes of objects including vehicles. We further fine-tuned the model with existing annotated datasets of developed country traffic from PASCAL VOC~\cite{pascalvoc} and KITTI~\cite{kitti2}. However, Mean Average Precision or MAP value in object detection was 0.01\% using fine-tuning with KITTI dataset and 0.58\% using fine-tuning with Pascal VOC dataset, on test video frames from Delhi-NCR. As this accuracy was not useful for any application, we build a parallel dataset in this paper annotating videos from roadside cameras in Delhi-NCR, to evaluate state of the art CNN methods in the developing region context. This dataset and our trained models have been made available, so that vision researchers can test their methods on this new dataset for developing countries, in addition to existing ones~\cite{trancos, kitti2, cityscapes} for developed countries. 

Non-laned heterogeneous traffic in developing regions has excited the research community to design automated traffic monitoring systems using a wide variety of embedded sensors like cameras~\cite{dev-traff7, dev-traff8, dev-traff4}, microphones~\cite{dev-traff1, dev-traff2} and RF~\cite{dev-traff3, dev-traff5}. All these works are on congestion estimation, that outputs the level of traffic density or the length of vehicle queue on a given road stretch. Our work gives a superset of these outputs. We give classified vehicle counts for a given road stretch, which in summation can give the traffic density on the roads. In dense traffic, the furthest vehicle object that we detect in a video frame will give the length of the traffic queue. Thus all the prior works' results can be derived from the results we present in this paper. We additionally report vehicle type, which as discussed in Section~\ref{sec-motivation}, has numerous applications of its own. The main technical gap compared to the related work has been in using the recent dramatic improvements in computer vision accuracy based on CNNs, not explored in prior literature. Vehicle classification was done using in-vehicle smartphone sensors in~\cite{dev-traff6}. This dependency on participation from on-road vehicles for vehicle classification has been removed in this paper, using data from roadside camera deployments.

There have been some very recent works on evaluating CNN inference performance on mobile and embedded systems~\cite{evaluation1, evaluation2}. These works discuss the image classification task, where a given image has a single object that needs to be classified into one of the pre-defined ground truth classes. We evaluate the more challenging multi-class object detection task, necessary to handle different applications on crowded traffic scenes from roadside cameras. Another point of distinction with the prior CNN evaluations on embedded platforms is their evaluations have been on high end platforms like Jetson TX2, which cost in the the order of 1000 USD. As we show by empirical evaluations, embedded platforms within 50-100 USD price ranges are also suitable for the use cases described in this paper,a promising observation to reduce deployment cost.
 
\begin{figure*}[t!]
\centering
\subfloat[Frontal view, fish eye lens]{\includegraphics[width=7cm, height=5cm]{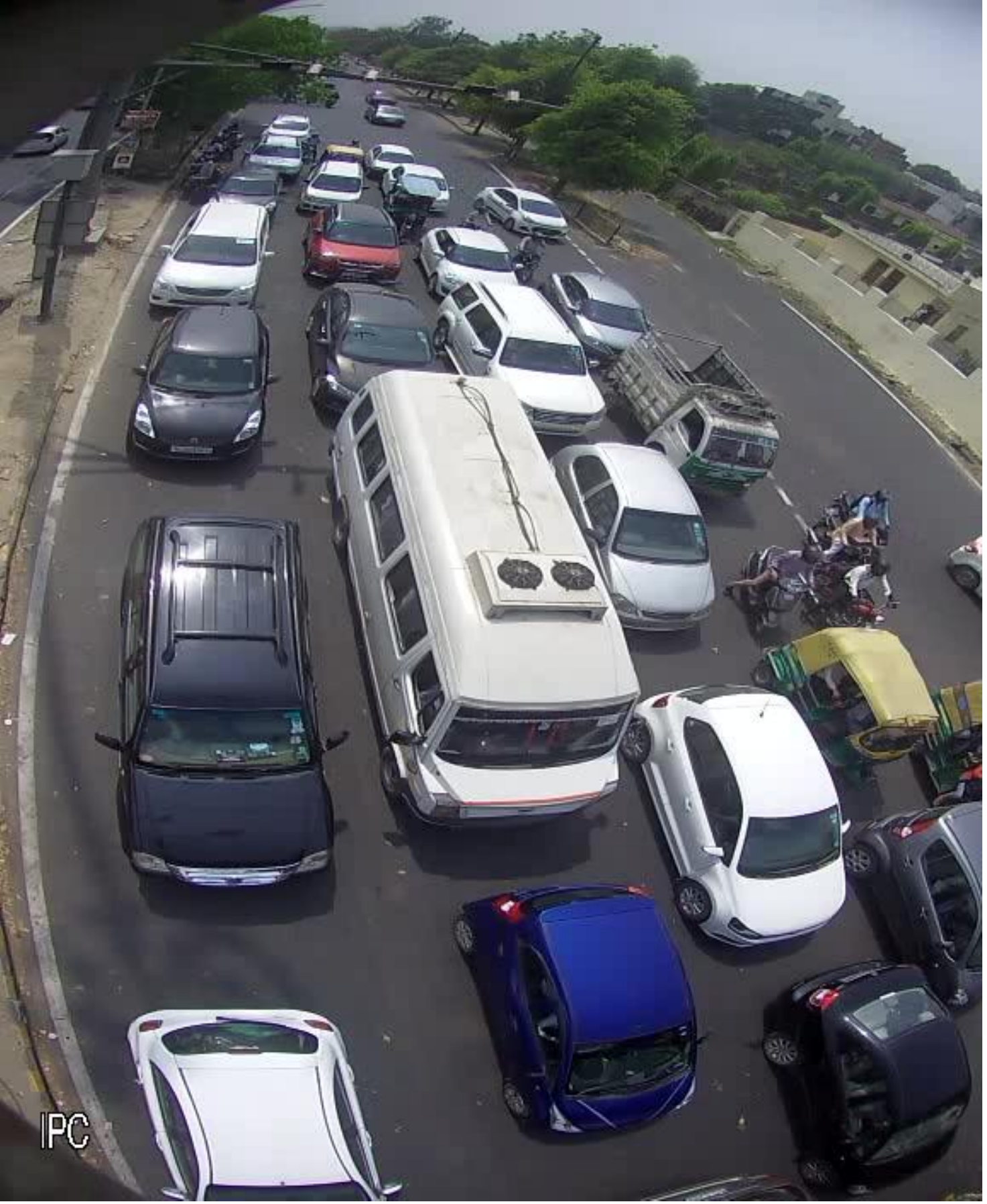}}\quad
\subfloat[Back view, non-bus route]{\includegraphics[width=7cm, height=5cm]{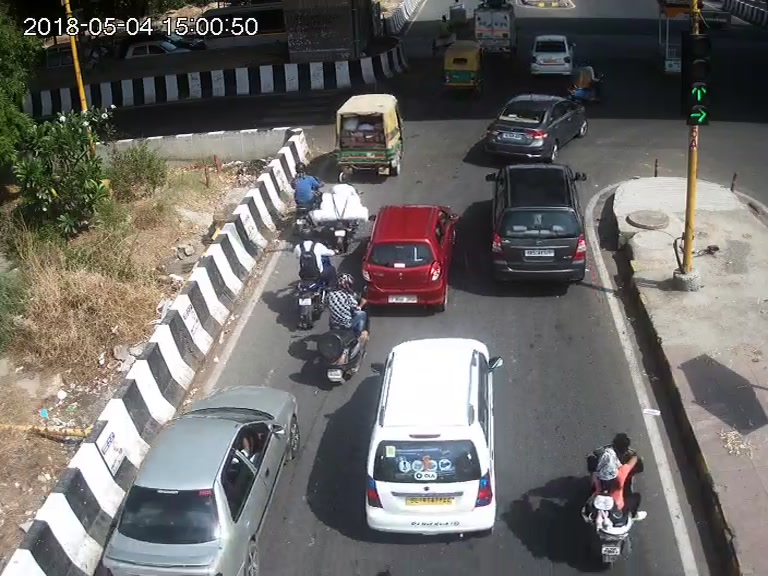}}\quad
\subfloat[Side view, highway]{\includegraphics[width=7cm, height=5cm]{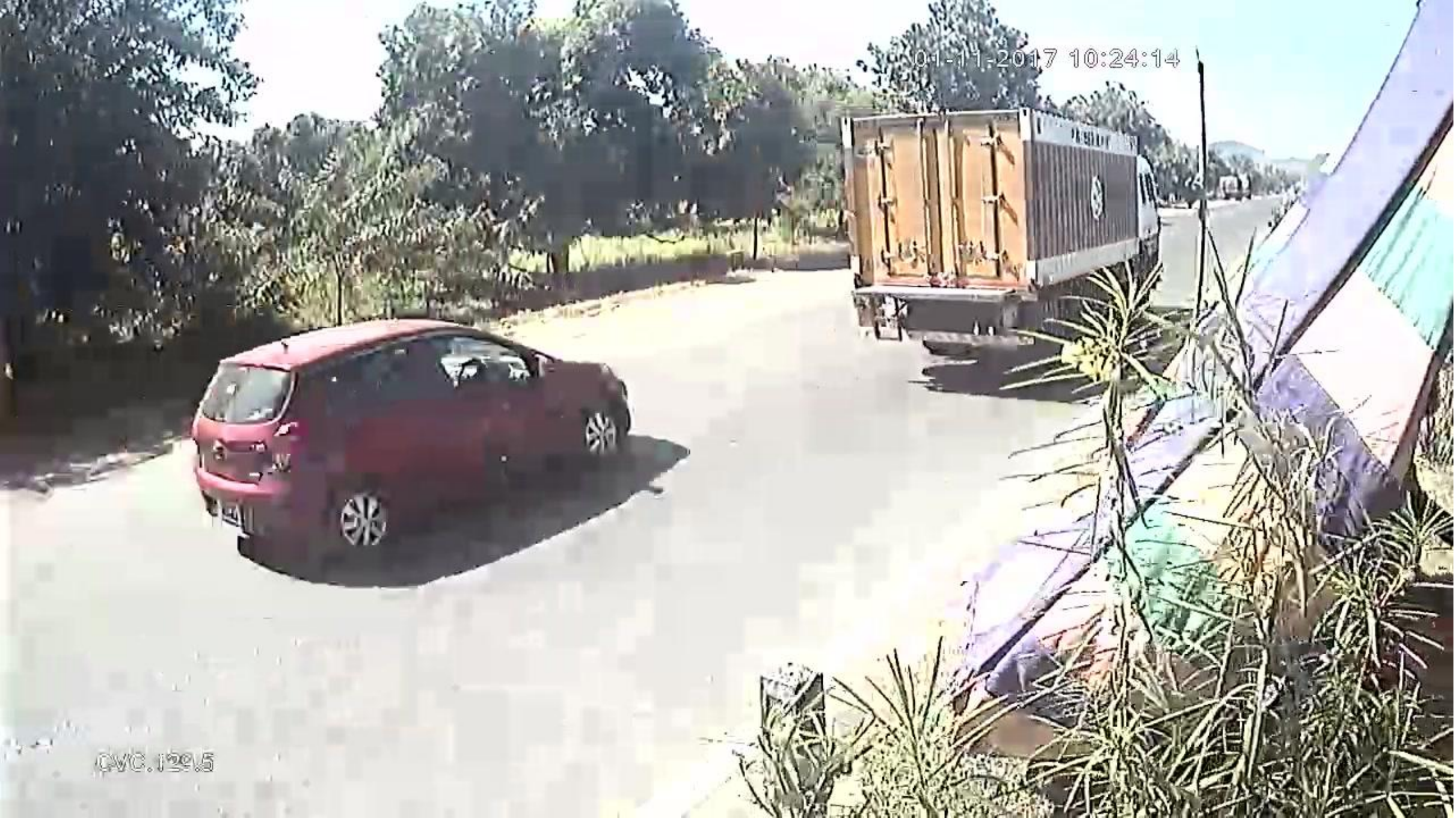}}\quad
\subfloat[Back view, bus route]{\includegraphics[width=7cm, height=5cm]{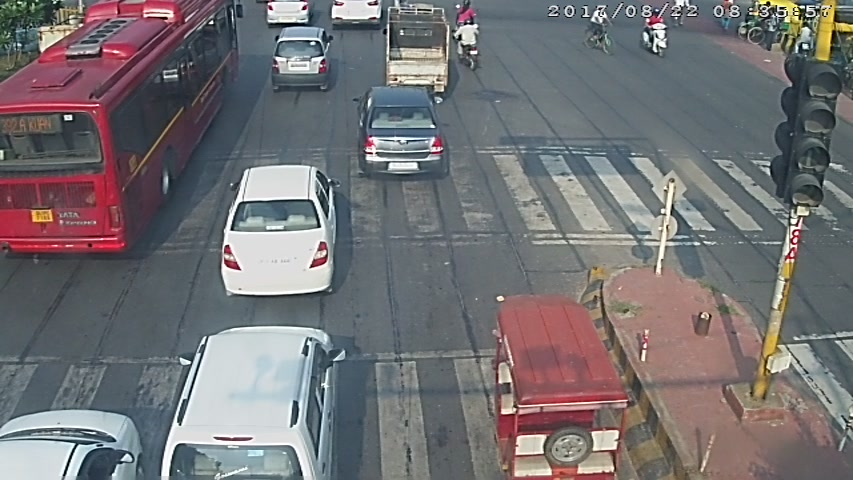}}
\caption{Four camera installations across Delhi-NCR}
\label{fig-installations}
\end{figure*}

\section{Dataset}
\label{sec-dataset}
We collected video frames from four roadside camera installations in collaboration with Vehant Technologies~\cite{vehant} and DIMTS~\cite{dimts}. As automated vehicle counting and classification is a new technology being tried in few areas of Delhi-NCR, the positioning and hardware specifications of the camera installations have not yet been standardized. Thus we have cameras showing frontal, back and side view of vehicles, and one camera even has a fish-eye lens. Sample images from the four installations are shown in Figure~\ref{fig-installations}. The difference between the two back view installations is in the type of vehicles that ply the monitored roads, one is a bus route where the other do not see buses. The datasets will be referred to as \one{}, \two{}, \three{}  and \four{}  henceforth. 

The first three rows in Table~\ref{tab-dataset} show the total number of annotated frames, the total and average number of annotations. We edit the open source BBox tool~\cite{bbox} to annotate images with rectangular bounding boxes and a label among one of the following 6 classes: 0-bus, 1-car, 2-autorickshaw, 3-twowheeler, 4-truck, 5-pedestrian, 6-cycles and e-rickshaws. Three of the authors annotated one-third of the images each. Then each annotator went through the annotations of the other two, and made small fixes. The annotations were thus by mutual agreement of three of the authors. The size of the annotated dataset is comparable to some of the existing datasets\footnote{The Trancos~\cite{trancos} dataset consists of 1244 images with a total of 46796 vehicles annotated}, and hence will be sufficiently big to be a useful benchmark for the computer vision community.

This annotated dataset is split into 80-20\% training and test sets respectively (last two rows in Table~\ref{tab-dataset}), in the next section where we train and evaluate our CNN models for the object detection task.

\begin{figure*}[t!]
\centering
\subfloat{\includegraphics[width=6.5cm, height=4.5cm]{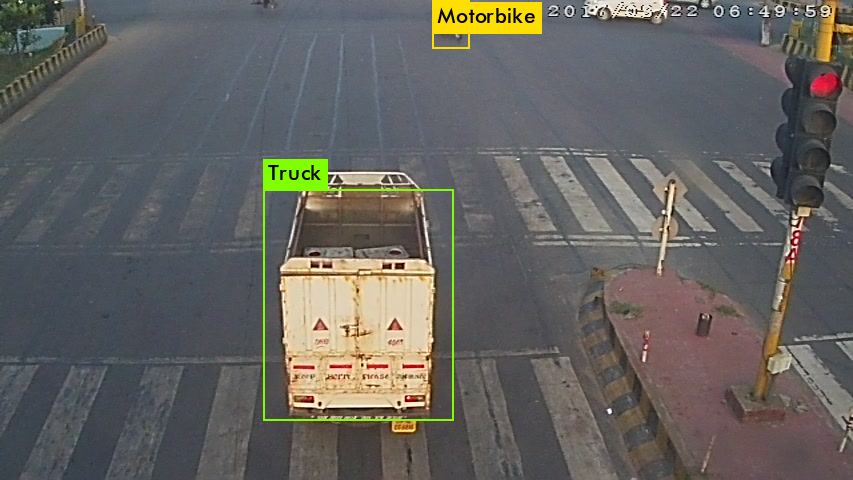}}\quad
\subfloat{\includegraphics[width=6.5cm, height=4.5cm]{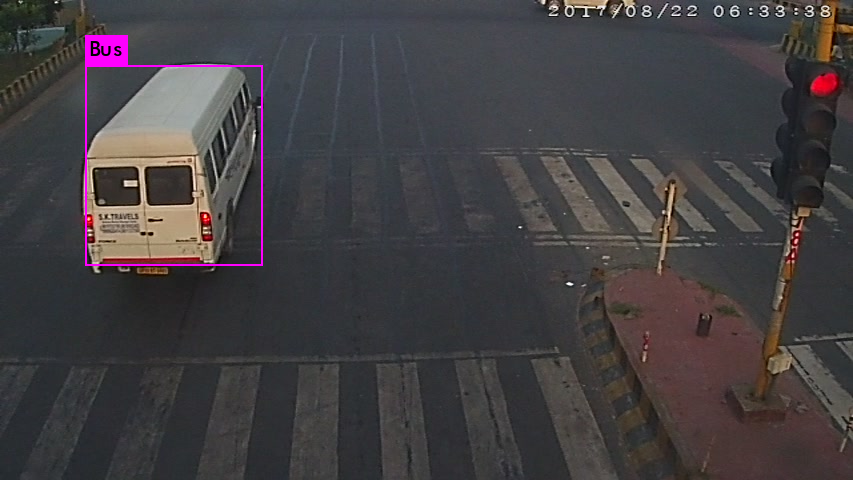}}\quad
\subfloat{\includegraphics[width=6.5cm, height=4.5cm]{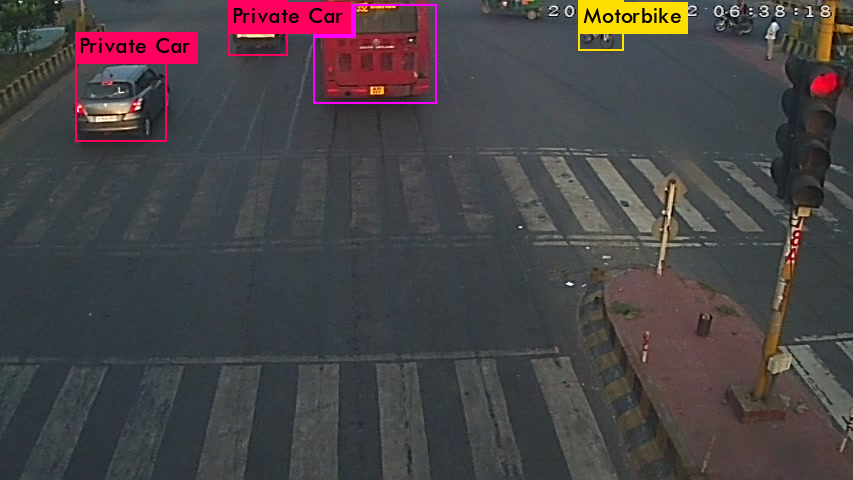}}\quad
\subfloat{\includegraphics[width=6.5cm, height=4.5cm]{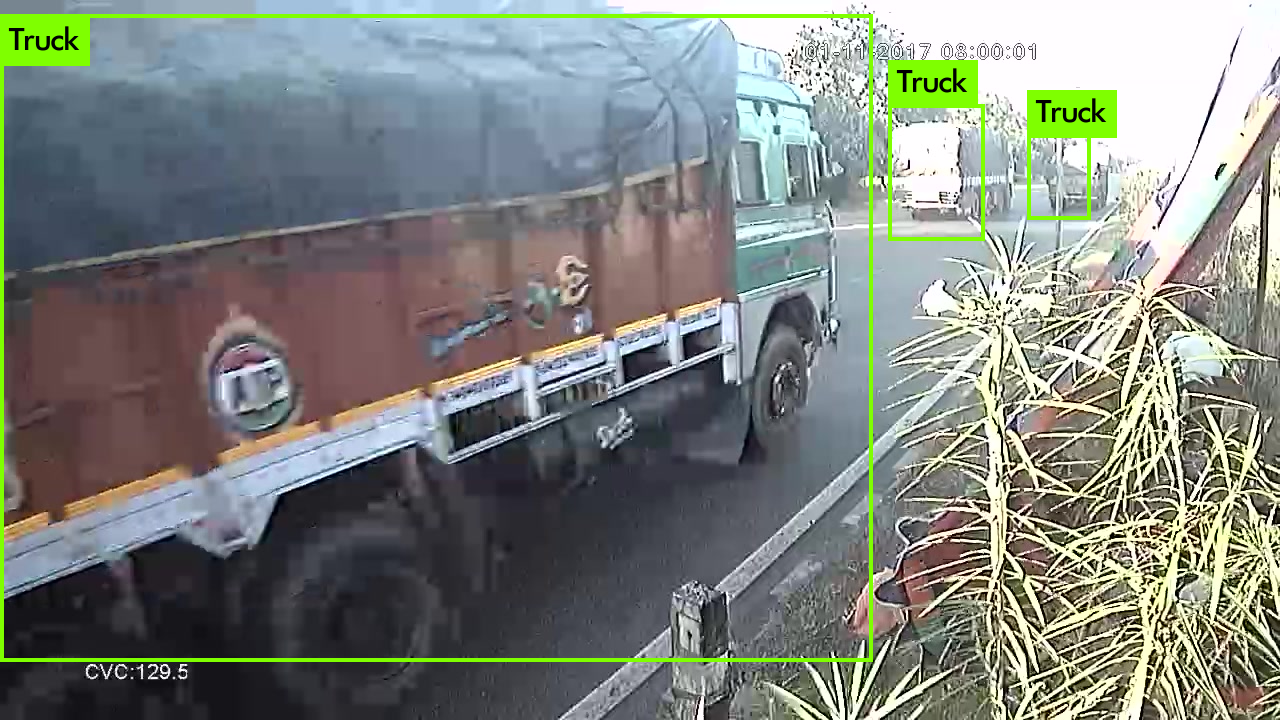}}\quad
\subfloat{\includegraphics[width=6.5cm, height=4.5cm]{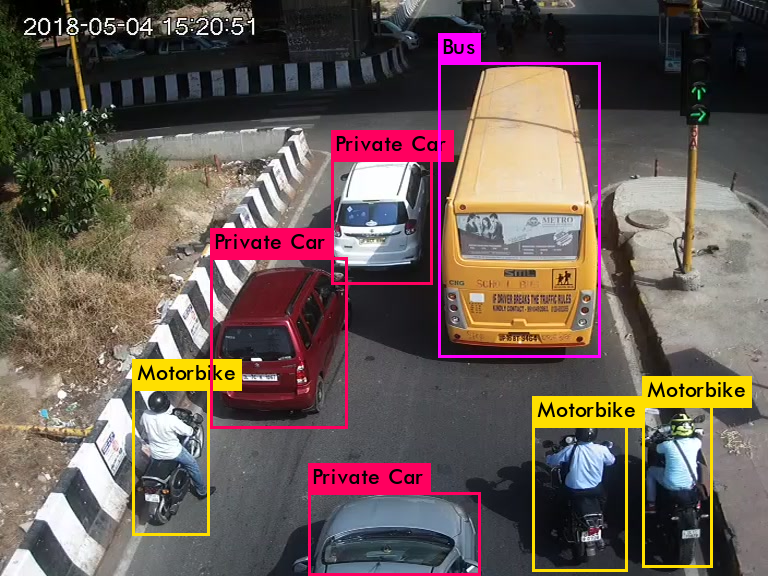}}\quad
\subfloat{\includegraphics[width=6.5cm, height=4.5cm]{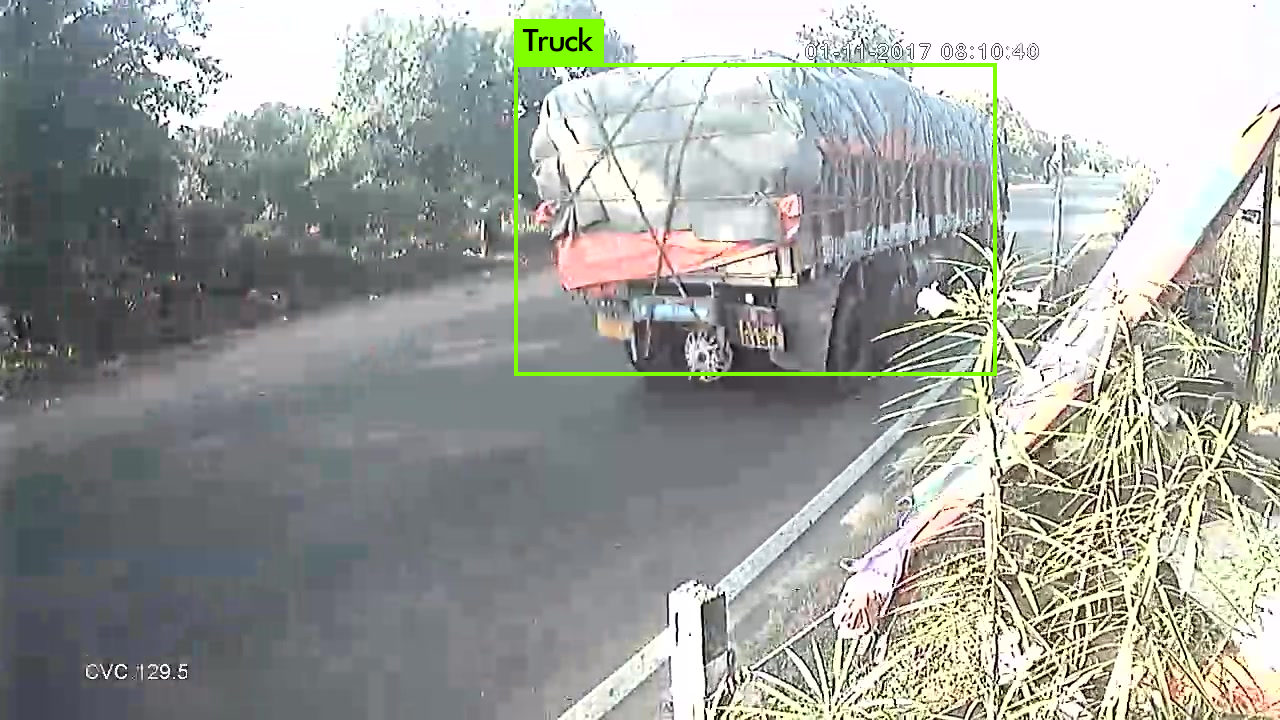}}\quad
\subfloat{\includegraphics[width=6.5cm, height=4.5cm]{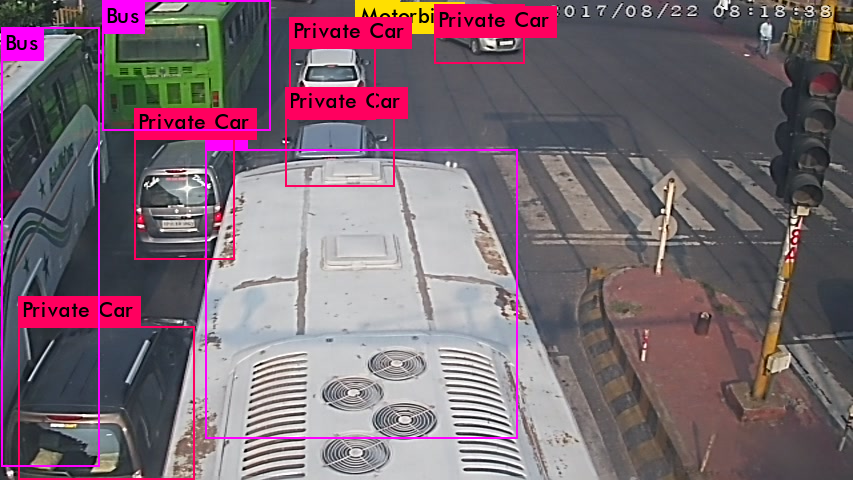}}\quad
\subfloat{\includegraphics[width=6.5cm, height=4.5cm]{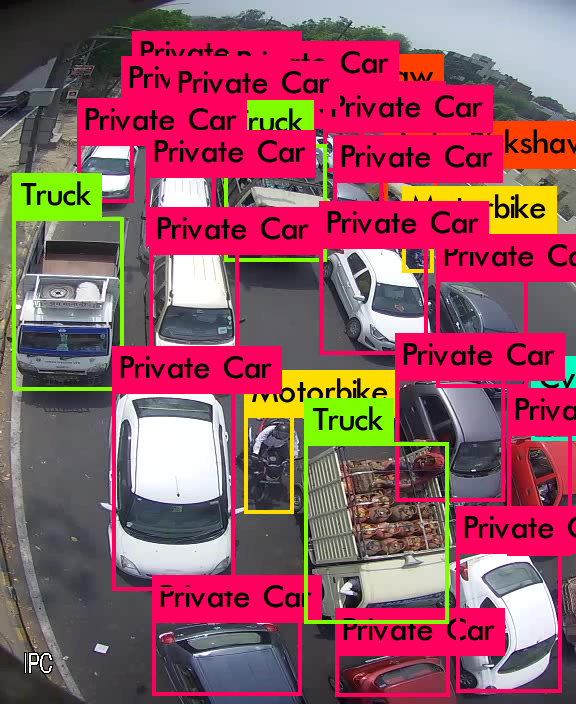}}\quad
\caption{Sample images with object detection bounding box and class label outputs}
\label{fig-detections}
\end{figure*} 

\section{CNN based object detection}
\label{sec-training}
For object detection and classification into our 7 annotation classes, we fine-tune the YOLO~\cite{yolo} CNN model, which has been known to give high object detection accuracy at low inference latency. In YOLO, a single Neural Network is applied to the whole image. The network divides the image into rectangular grid regions and predicts bounding boxes and probabilities for each region. Detections are thresholded by some probability value to only see high scoring detections (this process is known as non-maximum suppression in the computer vision community).

We use a model of YOLO pre-trained on the MS-COCO dataset. We fine-tune it on the PASCAL VOC 2007 and the KITTI datasets. This is followed by fine-tuning on our own custom annotated datasets (Table~\ref{tab-dataset} row 4). We fine-tune 5 different YOLO models, one model using exclusively the training data from a single camera installation. For four installations, this gives us four YOLO models. We fine-tune a fifth model, combining the training data from all installations. These models will be referred to as YOLO1, YOLO2, YOLO3, YOLO4 and YOLO5 in subsequent sections. Since each installation gives video frames of different resolutions, these frames are resized to 416 X 416 before being fed into the model for both fine-tuning and inference.

For the fine-tuning process, we use a Dell Precision Tower 5810 work-station, custom-fitted with an NVIDIA Quadro P5000 graphics card. For each model (except YOLO5), the weights from the $10600^{th}$ fine-tuning epoch is used to evaluate inference accuracy on the test data (Table~\ref{tab-dataset} row 5). For YOLO5, after fine-tuning for more than 80200 iterations, the best accuracies are obtained using weights from the $720006{th}$ iteration. Each training epoch takes about a second on this work-station, and therefore the fine-tuning process for the first four models take 3 hours and that for YOLO5 takes 20 hours.

\subsection{Object detection output visual examples}
Figure~\ref{fig-detections} shows the outputs of our detection models on a small set of example images. Each class is represented with a different colored box in the output. Before examining the accuracy of the object detection models rigorously using precision-recall numbers on the entire test dataset, these images give an idea of the excellent performance of the trained models for all four camera installation locations.

\begin{figure*}[t!]
\centering
\includegraphics[scale=0.7,angle=-90]{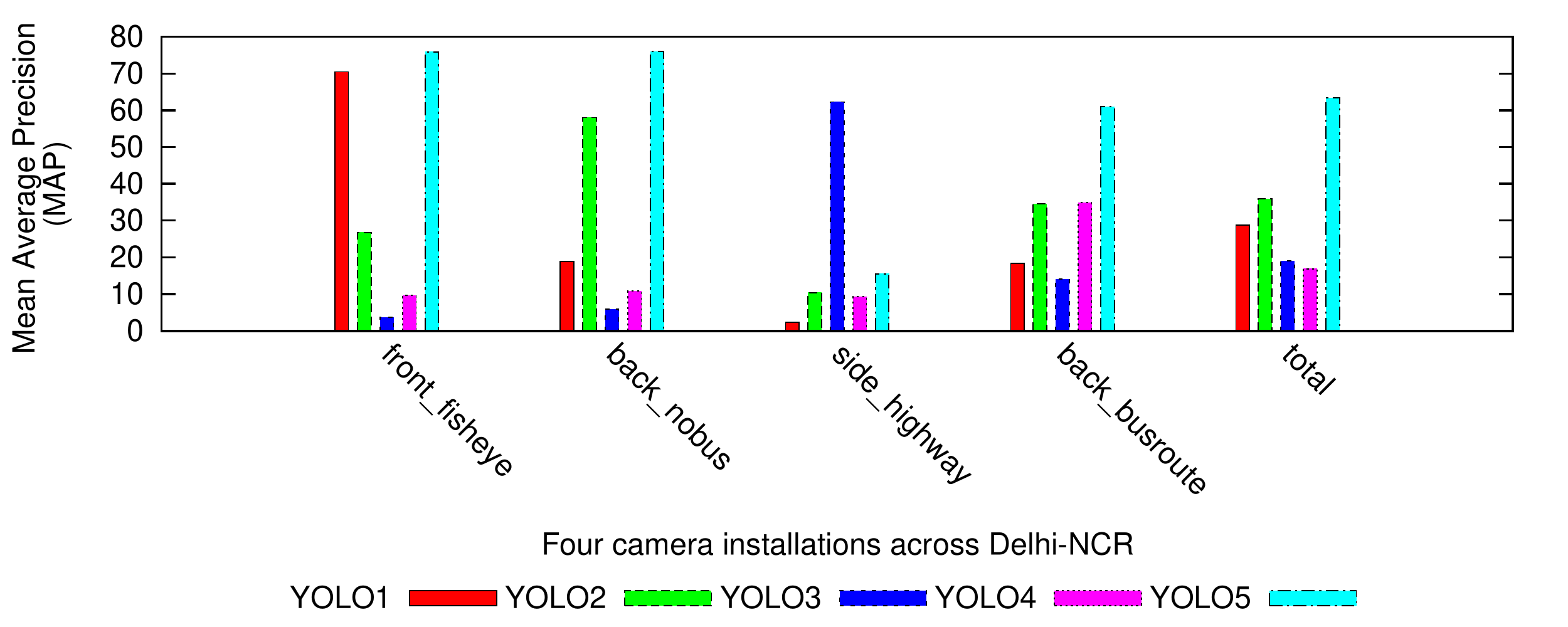}
\caption{Mean Average Precision (MAP) of the five object detection models on the five test sets (one from each of the four camera installations and a fifth combined test set)}
\label{fig-accuracy}
\end{figure*}
\begin{figure*}[ht!]
\centering
\subfloat[Precision for different object classes]{\includegraphics[scale=0.65,angle=-90]{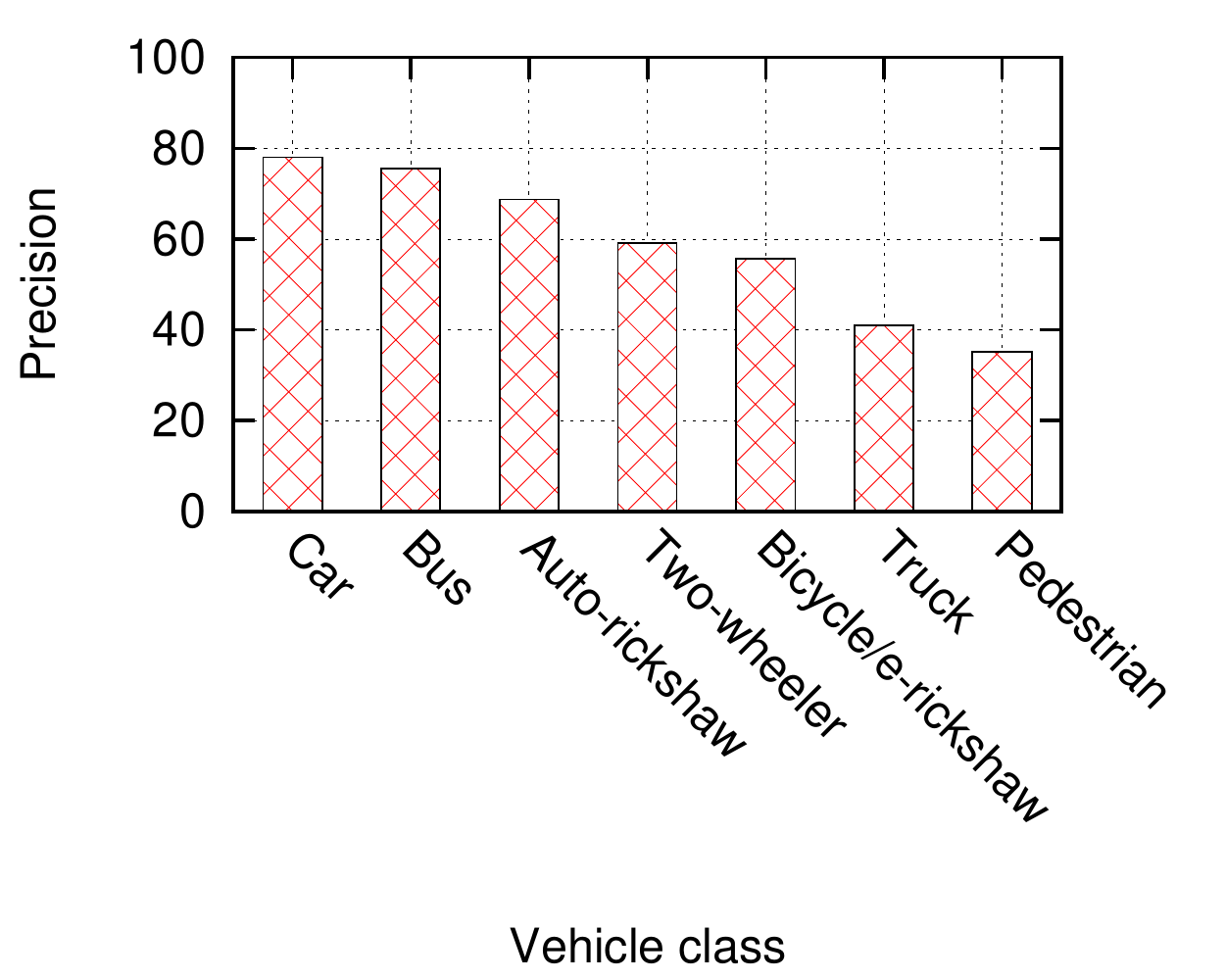}}\quad
\subfloat[Recall for different distances from the camera, the lower the y-pixel value the more is the distance from the camera as the top-left corner of the frame furthest from the camera has co-ordinate (0,0).]{\includegraphics[scale=0.65,angle=-90]{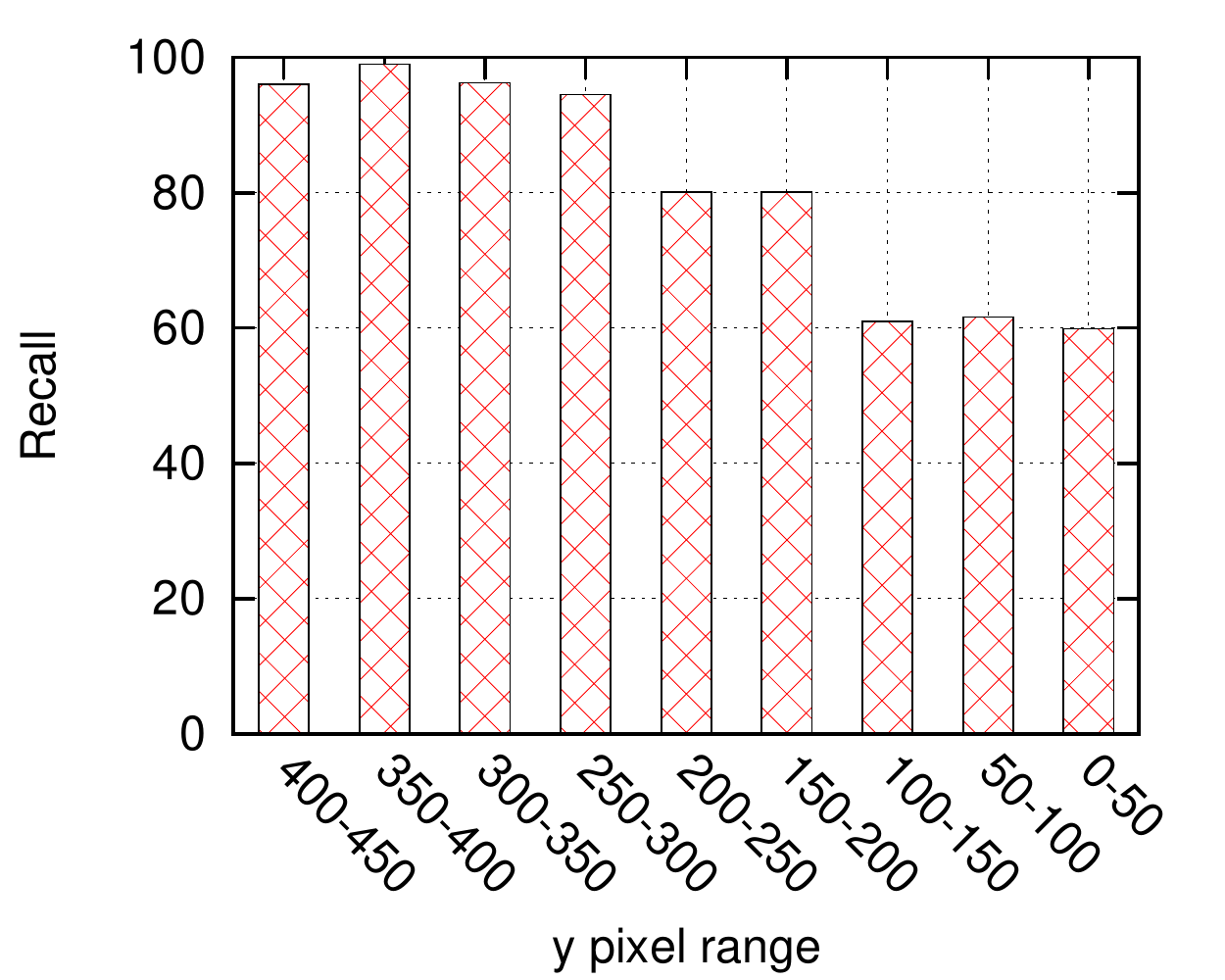}}
\caption{Object detection accuracy for different object classes and different object distances from the camera}
\label{fig-class-distance}
\end{figure*} 

\subsection{Installation specific fine-tuning}
As discussed above, we fine-tune five models (four fine-tuned using training data from each of the four camera installations and the fifth model trained with the combined training dataset). Evaluation is done individually with the five test sets (one from each of the four camera installations and a fifth combined test set). Then we run each of the five models on each of the five test sets, performing 25 evaluations in all. 

Mean Average Precision (MAP) values for these are plotted as bars along y-axis in Figure~\ref{fig-accuracy}. The first two locations \one{} and \two{} perform best with the combined model YOLO5, closely followed by YOLO1 and YOLO2 respectively, which are the models fine-tuned with training data from these two specific locations. Thus installation specific fine-tuning performs well in these locations, as well as combined model from all locations' training data. Similar pattern is seen for the fourth location \four{}, where the combined model YOLO5 again gives best results. YOLO4 (trained with this specific location's data) and YOLO2 give comparable accuracies in this location, the possible reason being both these models using back facing frames for training. 

The third location \three{} shows a distinct pattern. Here YOLO3, the model fine-tuned with this specific location performs well, but unlike in other locations, the combined model YOLO5 does poorly. The possible reason is this location having very distinct frames (side view with camera at the same level as the vehicles), as well as different kind of vehicles (mostly trucks as this is on a highway). Combining data from other locations with frontal or back views of different vehicle types reduce model accuracy in this location. YOLO5 however does well on the combined test set, that has test frames from all locations.

The main take-away from these evaluations is the necessity of standardizing camera hardware and its mounting for traffic applications. Manually annotating video frames for each distinct installation has a huge overhead. Installation specific models or combined models trained with annotated data from all installations will incur that overhead. However, if standardized, the trained models show promise of performing well on these vision tasks (as seen from the above 65-75\% MAP values using installation specific or combined models).

\subsection{Vehicle class specific accuracy}
Figure~\ref{fig-class-distance}(a) shows the vehicle class specific accuracies using YOLO5. On close examination, high intra-class variance and small number of training samples are found to reduce accuracy for some classes. Bicycles and e-rickshaws have huge intra-class variation. Similarly all kinds of lorries, trucks and smaller commercial vans have been labeled as trucks in our ground truth dataset, inducing large intra-class variance. In terms of small number of training samples, our videos being from busy intersections and highways, the number of pedestrians have been few in the dataset. Thus the class specific accuracy values for these classes of cycles/e-rickshaws, trucks and pedestrians have been low.

To increase accuracy for each object class, careful choice of labels (to reduce intra-class variations by having separate labels for very different looking objects in the same class) and enough training data per class, are recommended.  

\subsection{Recall vs. object distance from camera}
In addition to class specific inference accuracy, we also explore the dependency of accuracy on distance from the camera. Figure~\ref{fig-class-distance}(b) shows the distance of a ground truth object from the camera along x-axis, as y-pixel values in the video frame. With the top left corner of the frame as co-ordinate (0,0), smaller values indicate larger distance from the camera. Thus objects near the camera are very accurately detected, while accuracy drops further away from the camera.
\begin{table*}[t!]
\begin{center}
\begin{tabular}{ |c |c| c | c|}
\hline
Platform  &   Cost (INR) & Processor &  RAM\\ \hline
NVIDIA Jetson TX2 & 70K &  ARM Cortex-A57 (quad-core CPU) @ 2GHz + & 8 GB \\
                  &     &  NVIDIA Denver2 (dual-core CPU) @ 2GHz & \\ 
                  &     &  256-core Pascal GPU @ 1300MHz & \\\hline
Raspberry PI 3B   &  2.7K & Quad Core 1.2GHz Broadcom BCM2837 64bit CPU & 1 GB \\ \hline 
Raspberry PI 3B +    &  7.8K & Quad Core 1.2GHz Broadcom BCM2837 64bit CPU & 1 GB \\
Intel Movidius Neural Compute Stick &       &           Intel Movidius Vision Processing Unit & \\ \hline
\end{tabular}
\caption{Embedded Hardware Platforms}
\vspace{-0.7cm}
\label{tab-platforms}
\end{center}
\end{table*}
In each camera installation (front, back or side view), the same vehicles are in near and far field of the camera at different times. Thus application results (percentage of public vs. private transport or catching trucks if they ply outside their allotted hour) will not change if computations are restricted to the near field of the camera where object detection accuracy is high. This will reduce annotation overhead (not mark anything away by a certain distance from the camera) during training and increase accuracy (focused only in the near field of the camera) during inference. 

In summary, multi-class object detection for non-laned heterogeneous traffic on Delhi-NCR roads is fairly accurate (65-75\% MAP according to Figure~\ref{fig-accuracy}. This inference accuracy can be improved by better choice of labels to reduce intra-class variations, increasing training samples for each class and restricting computations within the near field of the camera (this last step increases recall to above 95\% as seen in Figure~\ref{fig-class-distance}(b)). Focusing computations to the near field of the camera also reduces object annotation overhead of far-away objects. Manual annotation overhead can be further reduced if camera hardware and installation directions and angles are standardized across all deployments, to remove the need of installation specific CNN model fine-tuning.    

\section{Embedded CNN inference}
\label{sec-embedded}
The training and evaluation of the CNN based object detection models in Section~\ref{sec-training} were done on a GPU server. For an on-road deployment of such a system, relying on fiber for transferring video from the road to the remote GPU server will be difficult in developing regions, as broadband connectivity across different road intersections might vary. Also, good cellular connectivity will incur recurring costs. 

This motivates us to explore the possibility of in-situ computer vision on embedded platforms. If using pre-trained CNN models (trained on GPU servers), state of the art CNN based object detectors can run their inference stage on embedded platforms co-located with the camera infrastructure on the roads, then only the counts per object class can be sent to the remote server for further analysis. The raw video streams need not be transmitted. We evaluate the cost, support for CNN software framework, latency and energy of multiple off-the-shelf embedded platforms in this section, for different CNN based object detection inference tasks. 

\subsection{Embedded Platforms}
Table~\ref{tab-platforms} lists the embedded platforms used in our evaluations. The first platform, NVIDIA Jetson TX2 is the most powerful embedded platform available in the market with impressive CPU and GPU support, and significant memory size. Its cost however is 10-24  times that of the other two platforms evaluated. The second platform, Raspberry PI has powerful CPU cores and moderate memory size, at a very  affordable price. The third platform, Intel Movidius Neural Compute Stick is an USB stick offered by Intel, which has specialized hardware called the Vision Processing Unit (VPU). Different computations necessary in vision tasks, like convolution operations in CNN, are implemented in hardware in this VPU. This hardware accelerator stick can be plugged into the USB port of a Raspberry PI, to create an embedded platform with boosted computer vision performance. Both Jetson TX2 and Raspberry PI run Linux based operating system like Ubuntu, on which different CNN software frameworks can be installed to run the object detection inferences given a trained model. We will refer to these three embedded platforms as jetson, raspi and movi respectively in subsequent discussion. 

We also evaluated two Android smartphones from Motorola and Samsung, but their cost-latency trade-offs in running the inference tasks were not comparable to these three platforms. Smartphones are more generic platforms targeted towards personal use, where cost increases due to the presence of different sensors, radios, display and also to support rich application software. The additional hardware/software are not necessary for dedicated tasks like on road traffic monitoring, hence embedded platforms with less features as listed in Table~\ref{tab-platforms} are more suitable. We therefore omit the evaluation numbers of smartphones from this discussion.

\begin{figure*}[t!]
\centering
\includegraphics[scale=0.7,angle=-90]{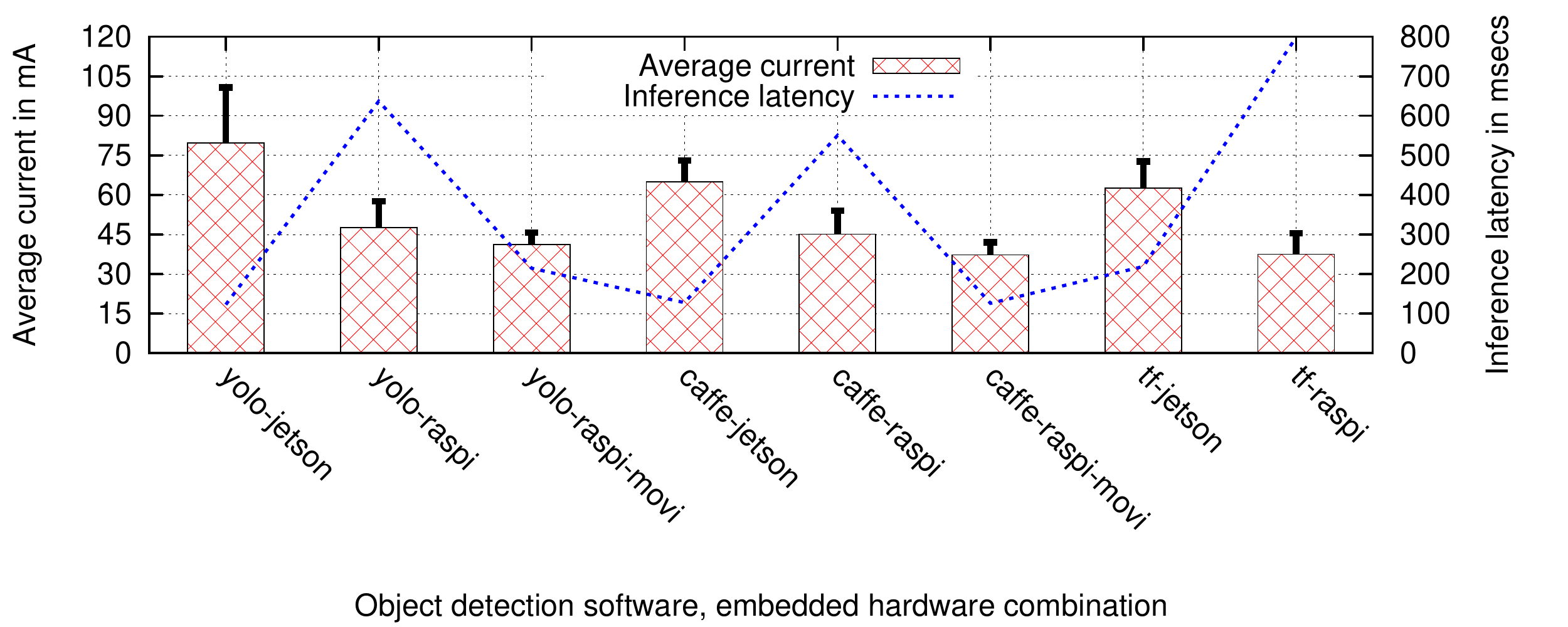}
\caption{Inference latency and energy of object detection DNN software on embedded hardware platforms}
\label{fig-latency-energy}
\end{figure*}

\subsection{CNN Software}
The explorations in Section~\ref{sec-training} used the YOLO object detector. Here we evaluate YOLO inference latency and energy on the embedded platforms. Additionally, we also evaluate the Mobilenet-SSD object detector, which has been reported to have similar accuracy and latency as YOLO. Just as we fine-tuned the YOLO object detection model using annotated video frames from our cameras, Mobilenet-SSD models can also be fine-tuned. To reduce the complexity of implementing every CNN from scratch, software frameworks like Tensorflow, Caffe, Pytorch are available, where functions for basic computational units like convolution, RELU, pooling etc. are already implemented. CNNs can be created calling these base functions as required. We use Mobilenet-SSD implementations on Caffe and Tensorflow. Thus we evaluate three CNN object detection software frameworks: YOLO, Tensorflow Mobilenet-SSD and Caffe Mobilenet-SSD. We will refer to these as yolo, tf and caffe  respectively in subsequent discussions. We download open source pre-trained models for these three softwares and only run the inference task on the three embedded platforms to measure latency and energy.

\subsection{Evaluation}
Figure~\ref{fig-latency-energy} shows the average current drawn with standard deviation as error bars on the left y-axis. Inference latency is shown on the right y-axis. The x-axis denotes different object detection software and embedded hardware combinations. The Movidius stick does not support running Tensorflow models, so the combination tf-raspi-movi is missing from the x-axis.

For a given CNN software (yolo, caffe or tf), there is a trend across the hardware platforms. Jetson uses high energy at low latency, raspi uses low energy with high latency and raspi-movi strikes an optimal balance using low energy comparable to raspi and incurring low latency comparable to jetson. For a given hardware platform (jetson, raspi or movi), there is again a trend among the CNN softwares. Yolo and caffe incur similar latency, while caffe incurs slightly less energy than yolo. Tf uses similar energy as caffe, but incurs higher latency than both yolo and caffe, especially on raspi. 

Given these trends, raspi-movi strikes a good balance of energy and latency in terms of hardware platform choice, at a moderate cost of 7.8K INR. This is significantly cheaper than the 70K INR jetson, while the jetson incurs similar latency and higher energy. In terms of software, both yolo and caffe Mobilenet-SSD are comparable in terms of latency and energy, while tf has higher latency especially on raspi and is currently not supported on raspi-movi.

While we evaluate and compare the different hardware platforms and software frameworks on the basis of latency and energy, depending on the application scenarios, one or both of these performance metrics might not be crucial. E.g. for an on road deployment, power is generally available from the lamp-posts or the traffic signals where the embedded computing units are deployed. For short term pilot deployments, not interfering with the road infrastructure and using battery supported units make sense, and energy efficiency will be important only in such scenarios. Similarly, if the goal of counting and classifying vehicles is to plan infrastructure or evaluate a transport policy like odd-even, low latency is not a necessity. Processing can be done at a slower rate, as no real time decisions will be taken based on the computer vision outputs. On the other hand, if the outputs are needed for real-time traffic rule violation detections and giving challans, then low latency becomes important. Thus the cost-latency-energy trade-offs of hardware-software combinations should always be considered in conjunction with the envisioned application requirements. Thus more important than the actual latency-energy values in our evaluation is the take-away, that state of the art CNN based object detectors can run on embedded platforms, thereby making in-situ processing of video frames feasible without depending on broadband connectivity. 

\section{Potential Applications}
\label{sec-applications}
We discussed many applications of classified vehicle counts in Section~\ref{sec-motivation}. Here we give a small example of how these applications can actually benefit from our system output. Table~\ref{tab-percentages} shows the percentages of different object classes in our dataset (excluding the highway dataset from \three{}). The numbers show a high dependency on private vehicles like cars and two-wheelers. This is the primary reason of increased traffic congestion and one of the potential factors in air quality degradation in Delhi-NCR. These values should be monitored when policies like odd-even are enforced to reduce congestion and air pollution, to check if public to private vehicle ratios improve, and whether absolute numbers of public vehicles like buses increase and cars come down.
\begin{table}[htb!]
\begin{center}
\begin{tabular}{ |c |c| c |}
\hline
Object class & Percentage on & Percentage on \\
&\four{} & \one{} and\\ 
& & \two{}\\\hline 
Bus & 3.03151 & 2.77197\\
Car & 52.3472 & 61.6214\\
Auto-rickshaw & 10.3994 & 8.42592\\
Two-wheeler & 26.5118 & 18.3368\\
Truck & 3.66009 & 4.47696\\
Pedestrian & 1.5993 & 2.23298\\
Cycle/E-rickshaw & 2.45067 & 2.13398\\\hline
\end{tabular}
\caption{Percentages of different vehicle classes in our dataset}
\label{tab-percentages}
\end{center}
\end{table}

The absolute counts of different vehicles is useful also to estimate the Passenger Count Unit (PCU)~\cite{pcu}, by multiplying the number of vehicles with the number of passengers each can carry. The PCU numbers will make infrastructure (signalized intersections, fly-overs, underpasses etc.) planning data-driven. Number of pedestrians can help in further planning of infrastructure like foot-bridges and side-walks. We will collaborate closely with the urban transport authorities to share our models for more extensive analyses across roads.

\section{Conclusion and Future work}
In this paper, we collect and annotate an extensive image dataset across four roads in Delhi-NCR. In future, we will enhance this labeled dataset with video frames from other Indian cities like Bengaluru and Mumbai and share these datasets and models with computer vision researchers and urban transport authorities. We achieve significant accuracy in classified object count using state of the art CNN models on non-laned heterogeneous traffic images. Performance on embedded platforms have also been shown to be practical in terms of latency and energy. Together, these form a very promising step towards building stronger collaborations with the traffic authorities, for scalable deployments of smart cameras and application design using our classified counts.

Such close collaboration is very important to understand the gap between Information and Communication Technology (ICT) and Development (D). Considering D, its clear that traffic congestion or road accidents have huge economic impact on developing economies. There is no paucity of such analysis highlighting the economic costs~\cite{congestion1, congestion2, congestion3, accident1, accident2}. On the ICT side, papers like this and others show the promise that traffic measurement, management and rule violation detection can be automated. How to bridge the gap between ICT and D, designing appropriate business models for such automation infrastructure, needs to be explored. In this context, the tension between better efficiency through automation and livelihood loss, is a burning question in developing regions. Whether manual data annotation to train automation models as used in this paper, manufacturing and packaging of embedded sensors, and occasional human supervision of the automated systems can balance the job requirements of the current traffic police force -- needs to be investigated. The authors' anecdotal discussions with the Delhi Traffic Police Commissioner suggests that they acknowledge the impossibility of manual management of continuously growing urban systems. The administration is actively looking for augmenting manual monitoring systems with automation, but high accuracy at low deployment and maintenance cost is needed. As immediate future work using the models trained in this paper, the authors are deploying a pilot network of connected intersection control cameras in Noida. The findings of installation and maintenance costs and effectiveness of the system in traffic measurement, management and rule violation detection and penalization will be shared with the traffic authorities.

As another follow-up work that will use the vehicle counting and classification models, built in this paper, is a project of the authors with the Delhi Integrated Multi Modal Transit System Limited (DIMTS). In this project, up to 300 DIMTS buses in Delhi-NCR will be instrumented with embedded sensors. GPS will measure the location of the bus, Particulate Matter sensors will measure PM 2.5 and PM 10 values, accelerometer will measure motion state of the bus as PM values captured in motion are noisy, temperature and humidity sensor values will be used to correct the PM readings and finally camera will be used to take pictures of road traffic and count and classify vehicles. These values will be collected across the city as the buses travel along their daily routes, stored locally and communicated via cellular radio to a central server. The classified vehicle counts will be correlated with PM values, to empirically measure potential and actual impact of policies like odd-even rules on air pollution. This project has been recommended under the DST-SERB IMPRINT II grant scheme, and will be undertaken as a joint project by IIT and IIIT Delhi and IIT Kanpur.

Finally, advocating for less cars to reduce congestion and air pollution is easy, but navigating the city might be difficult for the citizens if public transport facilities are not enhanced to match demand. Using Google Maps and Uber API, and also DIMTS bus mobile apps, the authors are trying to quantify the quality of public transport in Delhi-NCR using different metrics. Availability is the time to reach a public transport facility. Affordability is the cost of travel. Convenience captures different aspects of travel like number of breaks and total travel time in a trip, level of unpredictability in getting transport options etc. An automated set of tools to evaluate these metrics of public transport at city scale is being developed. While on-road camera based vehicle counting and classification gives one picture of road usage, web API based analyses give complementary information that together can give a coherent picture of the transport situation in the city.

Thus this paper is an intermediate point among a solid body of prior work~\cite{dev-traff1, dev-traff2, dev-traff3, dev-traff4, dev-traff5, dev-traff6, urbancomputing6} and an active line of future work by the authors to understand, quantify and hopefully better manage transportation in developing regions, with Delhi-NCR as a use case.

\balance
\bibliographystyle{ACM-Reference-Format}
\bibliography{references}
\end{document}